\documentclass{article}
\usepackage{spconf,amsmath,graphicx}
\usepackage{amssymb}
\usepackage{bm}

\usepackage{xcolor}
\usepackage{floatrow}

\usepackage[all]{kcompact}

\title{Unsupervised motion saliency map estimation \\ based on optical flow inpainting}
\twoauthors
 {L. Maczyta\sthanks{This work was supported in part by the DGA and the Région Bretagne as co-funding of Léo Maczyta’s PhD thesis}, P. Bouthemy}
{\normalsize Inria, Centre Rennes - Bretagne Atlantique, France}
 {O. Le Meur}
 {\normalsize Univ Rennes, CNRS, IRISA Rennes, France}

\begin{document}
%

\onecolumn
\section*{IEEE copyright notice}

\textcopyright~2019 IEEE. Personal use of this material is
permitted. Permission from IEEE must be obtained for all other uses,
in any current or future media, including reprinting/republishing this
material for advertising or promotional purposes, creating new
collective works, for resale or redistribution to servers or lists, or
reuse of any copyrighted component of this work in other works.\\
International Conference on Image Processing (ICIP) 2019.\\
DOI: 10.1109/ICIP.2019.8803542

\newpage

\twocolumn
\maketitle
\begin{abstract}
  
  The paper addresses the problem of motion saliency in videos, that
  is, identifying regions that undergo motion departing from its
  context. We propose a new unsupervised paradigm to compute motion
  saliency maps. The key ingredient is the flow inpainting
  stage. Candidate regions are determined from the optical flow
  boundaries. The residual flow in these regions is given by the
  difference between the optical flow and the flow inpainted from the
  surrounding areas. It provides the cue for motion saliency. The
  method is flexible and general by relying on motion information
  only. Experimental results on the DAVIS 2016 benchmark demonstrate
  that the method compares favourably with state-of-the-art video
  saliency methods.

\end{abstract}
\begin{keywords}

  Motion saliency, optical flow inpainting, video analysis

\end{keywords}
\section{Introduction}
\label{sec:intro}

Motion saliency map estimation corresponds to the task of estimating
saliency induced by motion. More specifically, regions whose motion
departs from the surrounding motion should be considered as
dynamically salient. Estimating motion saliency can be useful for a
number of applications, such as navigation of mobile robots or
autonomous vehicles, alert raising for video-surveillance, or
attention triggering for video analysis. In contrast to video saliency
approaches, we estimate motion saliency based on motion information
only. Indeed, we do not resort to any appearance cues to make the
method as general as possible. Furthermore, the method does not
require any supervised (or unsupervised) learning stage. Our main
contribution will consist in introducing flow inpainting to address in
an original way the motion saliency problem.

Video saliency has been first developed as an extension of image
saliency, with the objective to extract salient objects in
videos. Considering video means that temporal information becomes
available, and that motion is usable as an additional saliency cue. In
\cite{Wang2015}, Wang et al. rely on intra-frame boundary information
and contrast, as well as motion to predict video saliency. In
\cite{Le2016}, Le and Sugimoto propose a center-surround framework
with a hierarchical segmentation model. In \cite{Karimi2016}, Karimi
et al. exploit spatio-temporal cues and represent videos as
spatio-temporal graphs with the objective of minimizing a global
function.

Apparent motion in each frame is strongly influenced by camera
motion. While some approaches directly combine spatial and temporal
information without first cancelling the camera motion such as in
\cite{Fang2014, Kim2014, Mahapatra2014}, other methods explicitly
compensate the camera motion such as in \cite{LeMeur2007, Huang2014}.

Recently, deep learning methods have been explored to estimate
saliency in videos.  In \cite{Wang2018}, Wang et al. propose a CNN
exploiting explicitly the spatial and temporal dimensions, yet without
computing any optical flow. In \cite{Le2018}, Le and Sugimoto resort
to spatio-temporal deep features to predict dynamic saliency in
videos. They extend conditional random fields (CRF) to the temporal
domain, and they make use of a multi-scale segmentation strategy. In
\cite{Wang2_2015}, Wang et al. introduce saliency information as a
prior for the task of video object segmentation (VOS), by using
spatial edges and temporal motion boundaries as features.

\begin{figure*}[!thbp]
  \centering
  \includegraphics[width=\linewidth, height=3.35cm]{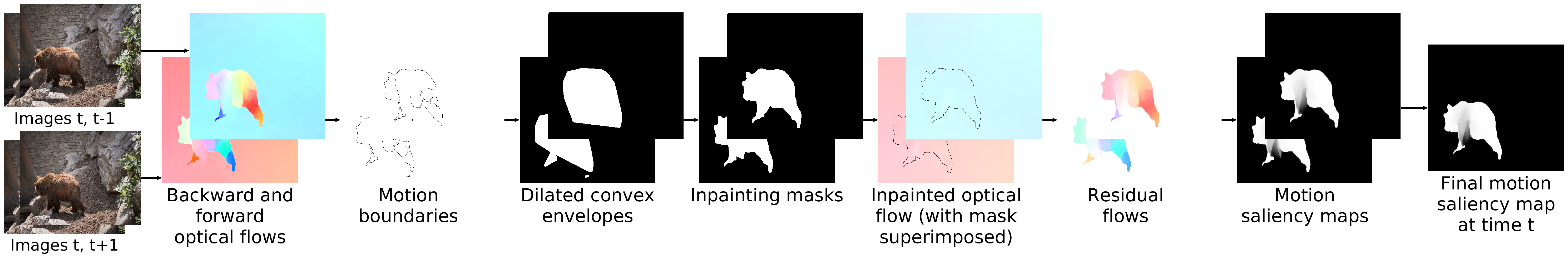}
  \caption{Overall framework of our method for motion saliency map
    estimation with the two backward and forward streams.}
  \label{figure/Overall framework}
\end{figure*}

The methods presented above are mostly directed toward the problem of
video saliency, that is, extracting foreground objects departing from
their context due to their appearance and motion. We are more
specifically concerned with the problem of motion saliency, which is
more general in some way, and highlights motion discrepancy
only. Configurations of interest may arise due to motion only, as in
crowd anomaly detection \cite{Perez2017b} (a person moving differently
than the surrounding crowd, or similarly an animal in a flock or a
herd, a car in the traffic, a cell in a tissue). In addition,
appearance can be of very limited use or even helpless for some types
of imagery, like thermal video or fluorescence cell microscopy.

The rest of the paper is organised as follows.
Section~\ref{section/Method} presents our method for motion saliency
estimation. Section~\ref{section/Experimental results} reports
comparative results with state-of-the-art methods for video saliency.
Section~\ref{Section/conclusion} contains concluding comments.

\section{Motion saliency estimation}
\label{section/Method}

As stated in the introduction, we estimate motion saliency maps in
video sequences only from optical flow cues. We expect that the
optical flow field will be distinguishable enough in salient regions.
We have to compare the flow field in a given area, likely to be a
salient moving element, with the flow field that would have been
induced in the very same area with the surrounding motion. The former
can be computed by any optical flow method. The latter is not directly
available, since it is not observed. Yet, it can be predicted by a
flow inpainting method. This is precisely the originality of our
motion saliency approach. Our method is then two-fold. We extract
candidate salient regions and compare the inpainted flow to the
original optical flow in these regions. A discrepancy between the two
flows is interpreted as an indicator of motion saliency. In addition,
we combine a backward and forward processing. Our overall framework is
illustrated in Figure \ref{figure/Overall framework}.

\subsection{Extraction of inpainting masks}

\begin{figure}
  \floatbox[{\capbeside\thisfloatsetup{capbesideposition={right,top},capbesidewidth=4cm}}]{figure}[\FBwidth]
  {\caption{Colour code (left) for the corresponding optical flow field (right).}\label{figure/hsv}}
  {  \begin{tabular}{cc}
       \includegraphics[width = 0.38\linewidth, height=0.38\linewidth]{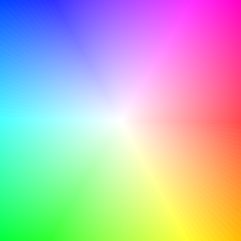}
       & \includegraphics[width = 0.38\linewidth, height=0.38\linewidth]{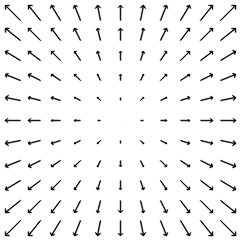}
     \end{tabular}
   }
\end{figure}

First, we have to extract the masks of the regions to inpaint. We will
rely on the optical flow field computed over the image, and more
precisely on its discontinuities. Indeed, the silhouette of any
salient moving element should correspond to motion boundaries, since
its motion should differ from the surrounding motion. The surrounding
motion will be generally given by the background motion, also referred
to as global motion in the sequel.

For the motion boundary extraction, one could directly apply a
threshold on the norm of the gradient of the velocity vectors. This is
however likely to produce noisy contours. Instead, we choose to rely
on the classical contour extraction method proposed by Canny
\cite{Canny1986}. For this, we convert the optical flow to its HSV
representation which is commonly used for visualisation, the hue
representing the direction of motion and the saturation its magnitude
(see Fig.\ref{figure/hsv}).

Then, we build region masks from these possibly fragmented contours as
illustrated in Figure \ref{figure/Overall framework}. Contours are
first organised into connected parts. For each connected part, the
convex envelope is computed and dilated with a 5x5 kernel.  Each final
region mask is given by the corresponding union of overlapping dilated
convex envelopes. By construction, region masks tend to be larger than
actual salient areas. Nevertheless, this is desirable for inpainting,
since inpainting must start from global motion information only. Yet,
a too rough mask can decrease the accuracy of motion inpainting,
especially for the salient areas which are non convex (see
Fig.\ref{figure/Overall framework}). The masks are then refined by
applying the GrabCut algorithm \cite{Rother2004} on the HSV
representation of the optical flow.  To avoid small localisation
errors which would include salient pixels in the inpainting mask, a
dilatation with a 5x5 kernel is again applied to the resulting mask.

\subsection{Optical flow inpainting}
\label{section/Saliency estimation}

We dispose of a set $\mathcal{R}$ of inpainting masks in the image
domain $ \Omega $. The issue now is to perform the flow inpainting in
these masks from the surrounding motion. We have investigated three
inpainting techniques to achieve it: two PDE-based methods
\cite{Telea2004, Bertalmio01} and a parametric method. Since the
background motion to inpaint is globally smooth, a diffusion-based
approach for inpainting is well-suited.

We apply the image inpainting method based on fast marching
\cite{Telea2004} as done in \cite{Strobel2014} for video completion,
which is a different goal than ours. We similarly extend the
Navier-Stokes based image inpainting method of \cite{Bertalmio01} to
flow inpainting.  We adopt the floating point representation of the
velocity vectors $\{\bm{\omega}(p), p \in \Omega\}$ with
${\bm\omega}(p) \in \mathbb{R}^2$.  The two components of the flow
vectors are inpainted separately.  Finally, we developed a parametric
alternative. We assume that the surrounding motion, i.e., the
background motion, can be approximated by a single affine motion
model. The latter is estimated by the robust multiresolution method
Motion2D \cite{Odobez1995}. The inpainting flow is then simply given
by the flow issued from the estimated affine motion model over the
masks. The three variants are respectively named MSI-fm, MSI-ns and
MSI-pm (MSI stands for Motion Saliency Inpainting).

\subsection{Motion saliency map computation}

The residual flow, i.e., the difference between the optical flow and
the inpainted flow, is then computed over the masks
$r, r \in \mathcal{R}$. The motion saliency map $g$, normalised within
$[0,1]$, is derived from the residual flow as follows:
\begin{equation}
  \label{equation/saliency}
  \forall p \in \mathcal{R}, \quad g(p) = 1 - \exp(-\lambda || \bm{\omega}_{inp}(p) - \bm{\omega}_{}(p) ||_{2}),
\end{equation}
\noindent where $ \bm{\omega}_{} $ is the optical flow, $ \bm{\omega}_{inp} $
the inpainted flow, $ g(p) = 0 $ for $p \notin \mathcal{R}$, and
$\lambda$ modulates the saliency score. Function $g$ expresses that
non-zero residual motion highlights salient moving elements. Parameter
$ \lambda $ allows us to establish a trade-off between robustness to
noise and ability to highlight small but still salient motions.

Let us note that, if we were interested in an explicit motion
segmentation, that is, producing binary maps, we would just need to
set $ \lambda $ to a high value. Indeed, by applying a threshold
$ \tau $ to $g(p)$, we can deduce from (\ref{equation/saliency}) that
 $ p $ will be segmented if:

\begin{equation}
  || \bm{\omega}_{inp}(p) - \bm{\omega}_{}(p) ||_2  \geqslant - \frac{ln(1-\tau)}{\lambda}.
\end{equation}

With $\tau$ arbitrarily set to $ \frac{1}{2} $ (the middle value of
$[0,1]$), the decision depends only on $\lambda$. Pixels with residual
flow magnitude greater than $\frac{ln(2)}{\lambda}$ will be
segmented. This shows that our method is flexible, since we can shift
from the motion saliency problem to the video segmentation problem
just by tuning parameter $\lambda$.

Finally, we propose to further leverage the temporal dimension to
reduce the number of false positive, in particular close to motion
boundaries. To do this, we introduce a bidirectional processing (see
Fig. \ref{figure/Overall framework}). The whole workflow is applied
twice in parallel, backward and forward, that is, to the image pair
$I(t), I(t-1)$ and the image pair $I(t), I(t+1)$. This yields two
motion saliency maps, which we combine by taking their pixel-wise
minimum. The reported experimental results for our main method and the
NM method introduced in Section~\ref{section/Quantitative comparison},
will include this bidirectional processing.

\section{Experimental results}
\label{section/Experimental results}

\subsection{Experimental setting}

For the computation of the optical flow, we employ FlowNet 2.0
\cite{Ilg2017}. This algorithm can run almost in real time and
estimates sharp motion boundaries. This is important for the
successful extraction of inpainting masks.

For all the experiments, the parameters are set as follows. The Canny
edge detector is applied to the image smoothed with a Gaussian filter
of standard deviation $\sigma = 5$. The two thresholds for the Canny
edge detector are set to 20 and 60 respectively. For the inpainting
algorithm, a radius of 5 pixels around the region to inpaint is used.
Finally, the parameter $\lambda$ for the computation of the saliency
map has been set to $\frac{3}{2}$.

\begin{table*}[!htb]
  \begin{center}
    \begin{tabular}{|c||c|c|c|c|c|c|c|c|c|c|c|c|}
      \hline
      
      Method            & STCRF \cite{Le2018} & MSI-ns             & MSI-pm & MSI-fm &                             VSFCN \cite{Wang2018} & RST \cite{Le2016}   & LGFOGR \cite{Wang2015} & SAG \cite{Wang2_2015}  & NM \\ \hline \hline
      
      MAE $\downarrow$  & \textbf{0.033}      & \underline{0.043}  & 0.044  & 0.045                        &  0.055                 & 0.077               & 0.102                  & 0.103                  & 0.453     \\ \hline
      
      F-Adap $\uparrow$ & \textbf{0.803}      & \underline{0.735}  & 0.724  & 0.716                        &  0.698                 & 0.627               & 0.537                  & 0.494                  & 0.367     \\ \hline
      
      F-Max $\uparrow$  & \textbf{0.816}      & \underline{0.751}  & 0.750  & 0.747                        &  0.745                 & 0.645               & 0.601                  & 0.548                  & 0.612     \\ \hline \hline

      Appearance        & \textit{Yes}        & \textit{No}     &\textit{No} & \textit{No}                  & \textit{Yes}           &\textit{Yes}                  & \textit{Yes}                    & \textit{Yes}                    & \textit{No}        \\ \hline

      Motion            & \textit{Yes}        & \textit{Yes}   &\textit{Yes}  &\textit{Yes}                                 & \textit{Yes}           &\textit{Yes}                  & \textit{Yes}                    & \textit{Yes}                    & \textit{Yes}       \\ \hline
 
      Supervised        & \textit{Yes}        & \textit{No}     &\textit{No}   &\textit{No}                                  & \textit{Yes}           &\textit{No}                   & \textit{No}                     & \textit{No}                      & \textit{No}        \\ \hline
      
    \end{tabular}
    \caption{Comparison with state-of-the-art methods for saliency map
      estimation on the test set of DAVIS 2016. In bold, the best
      performance; underlined, the second best.  We also indicate
      whether the method relies on appearance information, on motion
      information, and whether it is supervised.}
    \label{table/comparison}
  \end{center}
\end{table*}

There is no available benchmark dedicated to motion
saliency. Therefore, we choose the DAVIS 2016 dataset for the
evaluation of our method. This dataset has been initially introduced
in \cite{Perazzi2016} for the video object segmentation (VOS) task.
It has also been recently used to evaluate methods estimating saliency
maps in videos, as in \cite{Wang2018, Le2018}. For the VOS task, the
object to segment is a foreground salient object of the video, which
has a distinctive motion compared to the rest of the scene. It makes
this dataset exploitable for motion saliency estimation, although
appearance plays a role.

\begin{figure}[!htb]
  \begin{center}
    \begin{tabular}{cccc}
      \includegraphics[height=0.140\textwidth, width=0.200\textwidth]{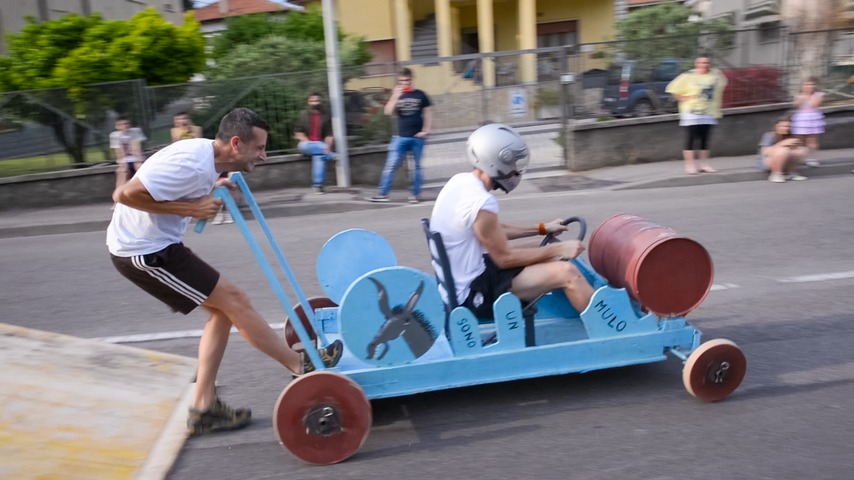}
      & \includegraphics[height=0.140\textwidth, width=0.200\textwidth]{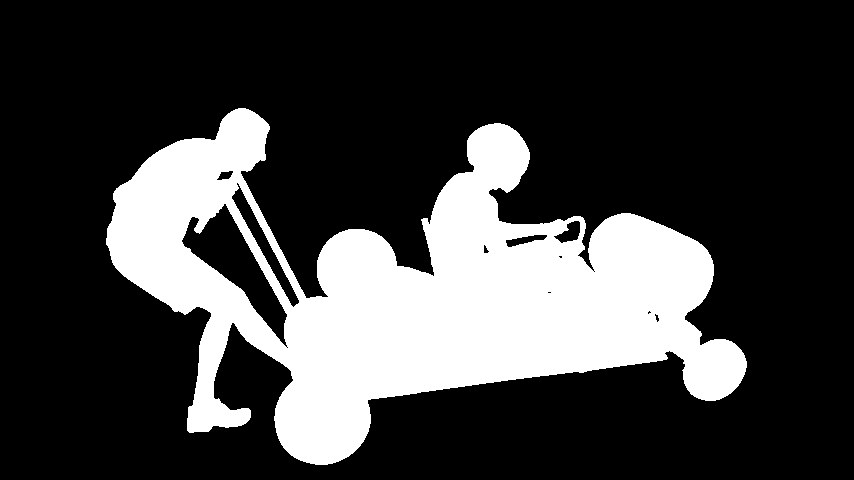}
      & \includegraphics[height=0.140\textwidth, width=0.200\textwidth]{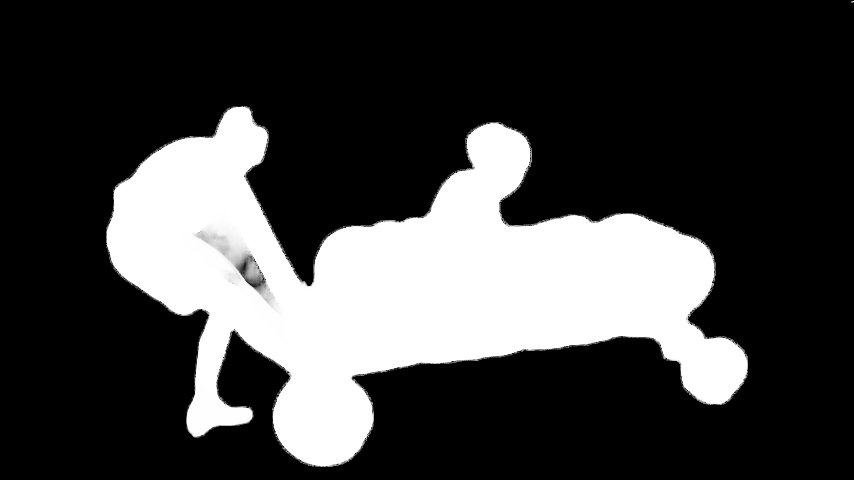}
      & \frame{\includegraphics[height=0.140\textwidth, width=0.200\textwidth]{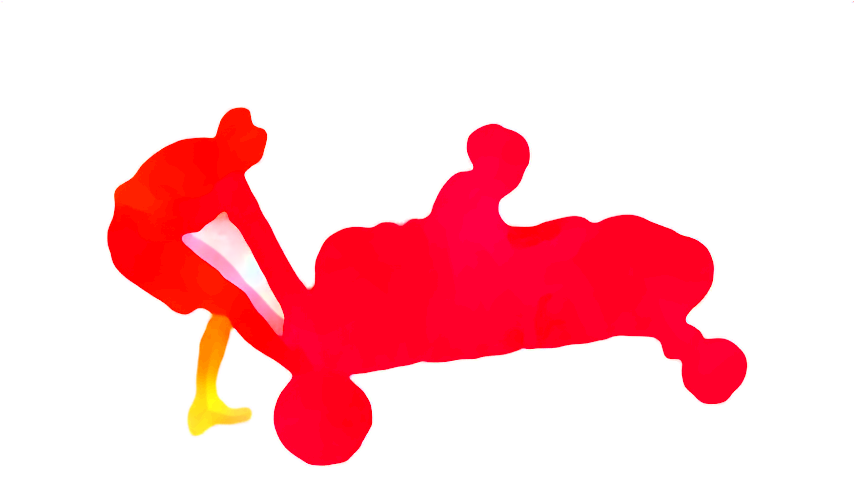}} \\
      
      \includegraphics[height=0.140\textwidth, width=0.200\textwidth]{Images/00003.jpg} 
      & \includegraphics[height=0.140\textwidth, width=0.200\textwidth]{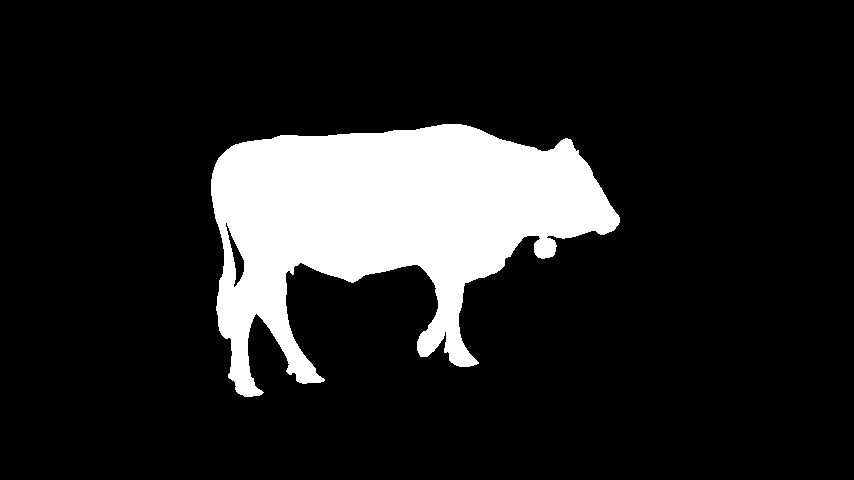}
      & \includegraphics[height=0.140\textwidth, width=0.200\textwidth]{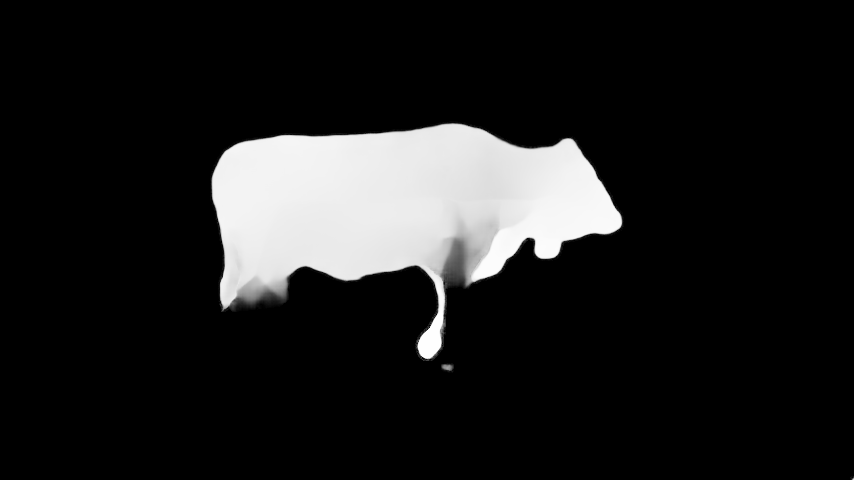}
      & \frame{\includegraphics[height=0.140\textwidth, width=0.200\textwidth]{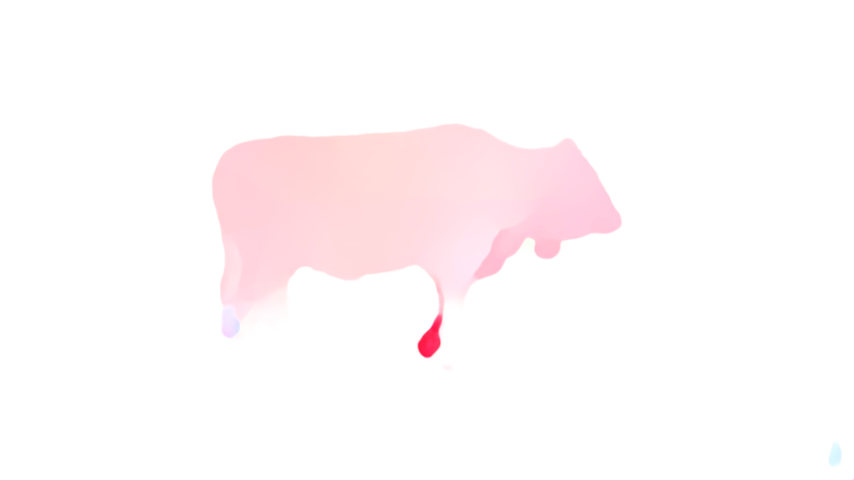}} \\
      
      \includegraphics[height=0.140\textwidth, width=0.200\textwidth]{Images/00037.jpg} 
      & \includegraphics[height=0.140\textwidth, width=0.200\textwidth]{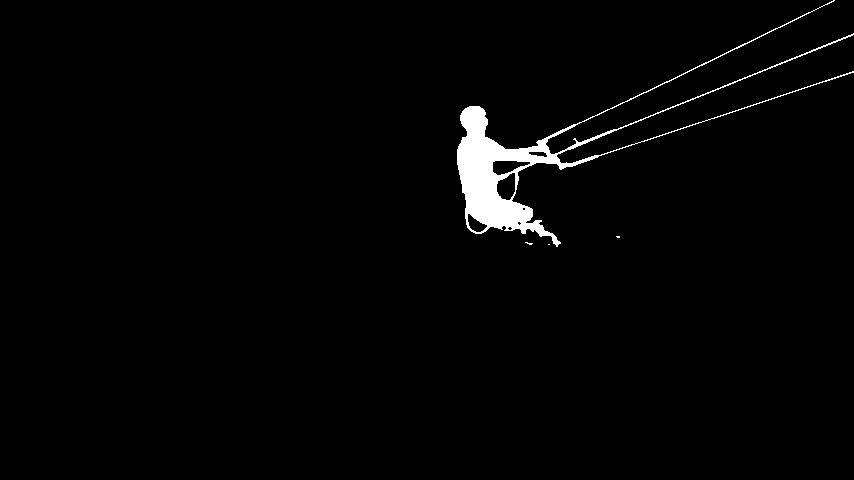}
      & \includegraphics[height=0.140\textwidth, width=0.200\textwidth]{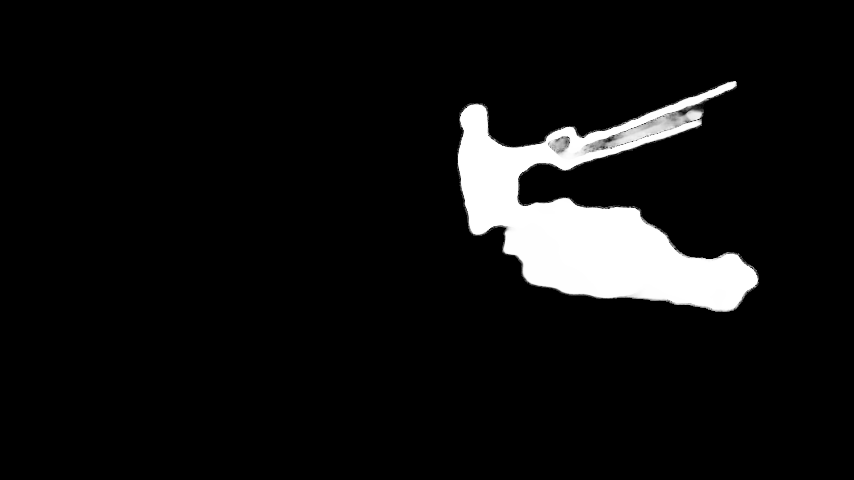}
      & \frame{\includegraphics[height=0.140\textwidth, width=0.200\textwidth]{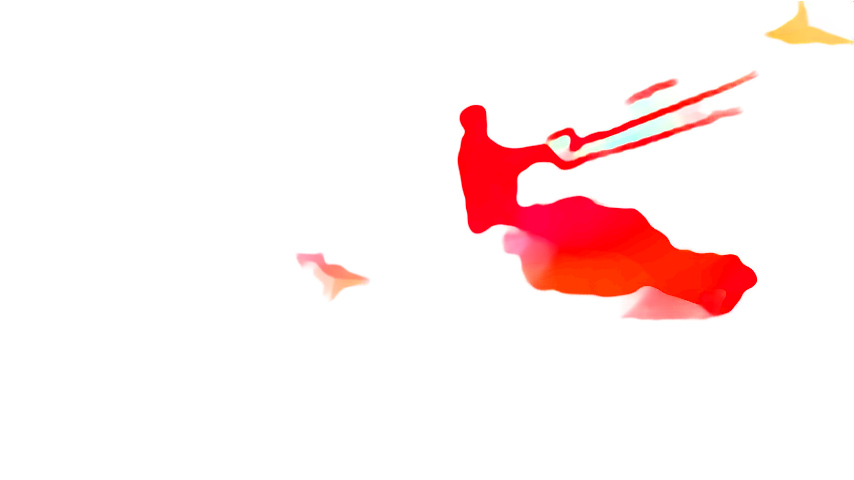}} \\
      
      \includegraphics[height=0.140\textwidth, width=0.200\textwidth]{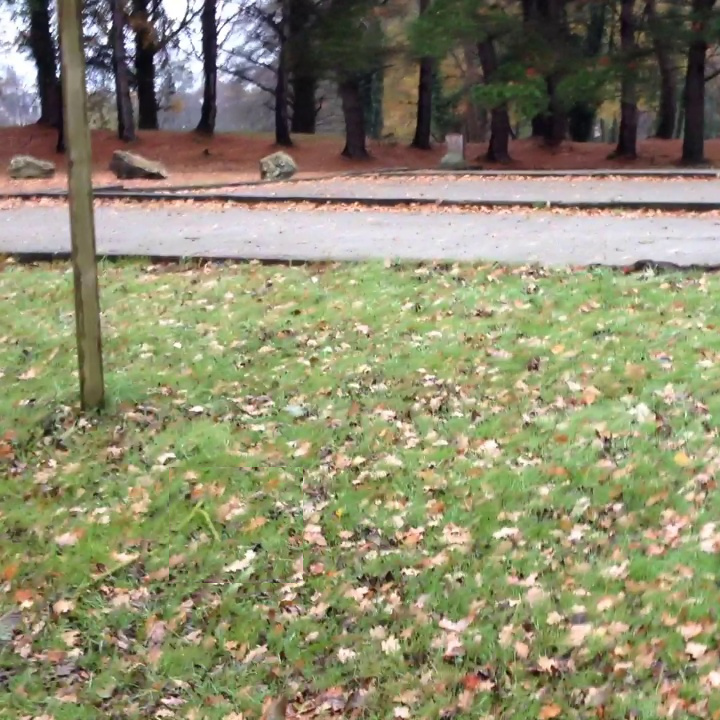} 
      & \includegraphics[height=0.140\textwidth, width=0.200\textwidth]{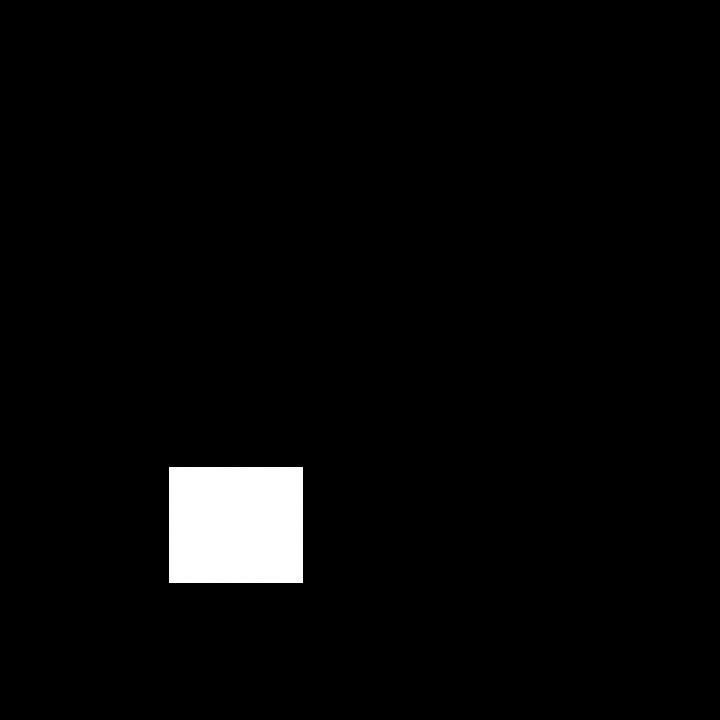}
      & \includegraphics[height=0.140\textwidth, width=0.200\textwidth]{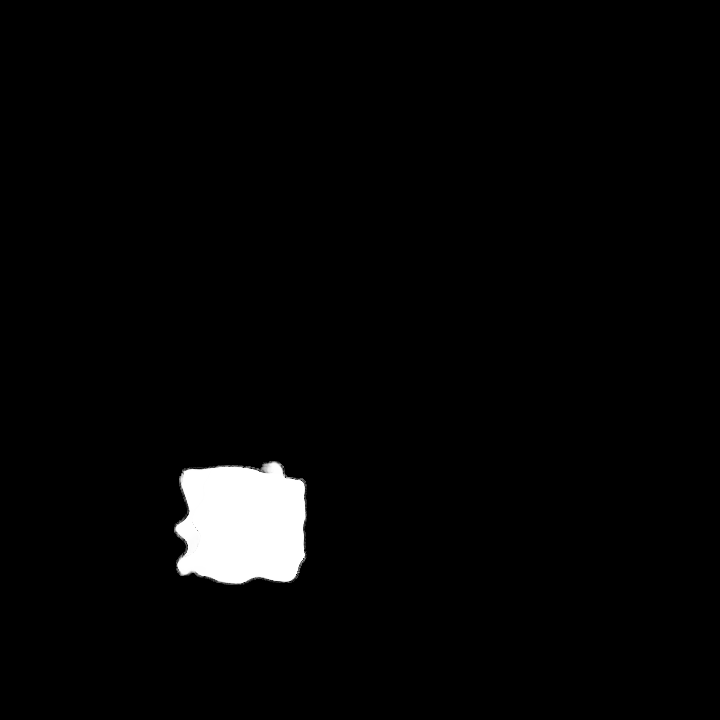}
      & \frame{\includegraphics[height=0.140\textwidth, width=0.200\textwidth]{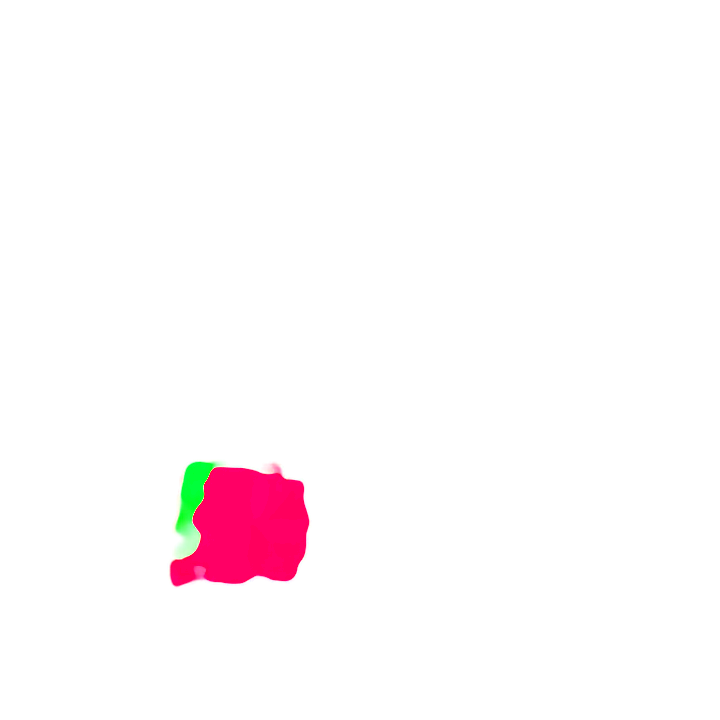}} \\
        
      \includegraphics[height=0.140\textwidth, width=0.200\textwidth]{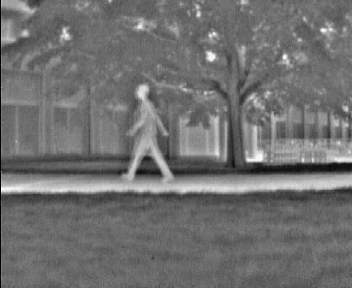} 
      & \includegraphics[height=0.140\textwidth, width=0.200\textwidth]{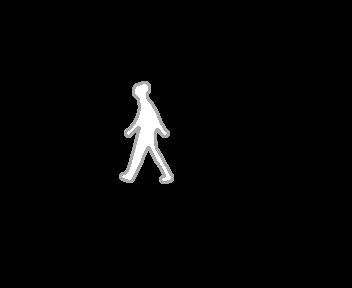}
      & \includegraphics[height=0.140\textwidth, width=0.200\textwidth]{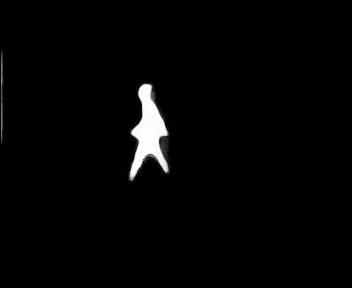}
      & \frame{\includegraphics[height=0.140\textwidth, width=0.200\textwidth]{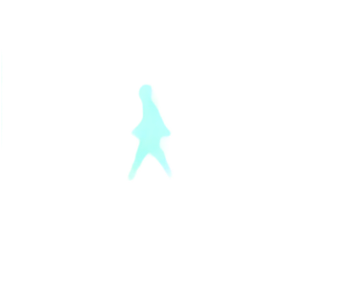}}
        
    \end{tabular}
  \end{center}
  \caption{\normalsize From left to right: one image from the video,
    binary ground truth, motion saliency maps predicted by our method
    MSI-ns, and the estimated forward residual flow (displayed with
    the motion colour code of Fig. \ref{figure/hsv}).}
  \label{figure/visualisation results}
\end{figure}

\subsection{Qualitative evaluation}
\label{section/Qualitative evaluation}

First, we present a visual evaluation of our method MSI-ns, which
turns out to be the best of the three variants as reported in
Table~\ref{table/comparison}. Fig. \ref{figure/visualisation results}
displays the output of our method for frames of the videos
\textit{soapbox, cows} and \textit{kite-surf} of the DAVIS 2016
dataset and for two other types of videos. In the fourth example
(\textit{lawn} video), a rectangular region in the lawn was
artificially moved in the image as indicated by the ground truth. It
provides us with an example where the only discriminative information
is supplied by the undergone motion. The fifth image comes from the
\textit{park} video of the changedetection.net dataset
\cite{Goyette2012}. It was acquired with a thermal camera, providing
us with an example where appearance is of limited help.

Both computed motion saliency maps and residual flows are shown in
Fig.~\ref{figure/visualisation results}. Indeed, the residual flow,
although an intermediate step in our method, is meaningful on its
own. It provides valuable additional information about the direction
and magnitude of salient motions in the scene. It could be viewed as
an augmented saliency map.

For the \textit{soapbox} example, the salient element with clearly
distinctive motion has been almost perfectly extracted. The
\textit{cows} example exhibits an interesting behaviour. The cow is
globally moving, except for its legs which are intermittently
static. This illustrates the difference between the video object
segmentation task, for which the whole cow should be segmented, and
the motion saliency estimation task, for which the elements of
interest are elements with distinctive motion. Our method consistently
does not involve the two legs in the saliency map.

In the \textit{kite-surf} example, the sea foam has a non rigid but
strong motion, and consequently, it is likely to belong to the salient
moving region, whereas for the VOS task, the kite-surfer is the only
foreground object to segment as defined in the ground truth.

In the \textit{lawn} example, the square region is easy to detect when
seeing the video, but is much harder to localize in a single frozen
image. Our method based on optical flow is able to recover the salient
moving region. Finally, in the \textit{park} example involving an IR
video with less pronounced appearance, our method also yields a
correct motion saliency map.

\subsection{Quantitative comparison}
\label{section/Quantitative comparison}

We introduce a naive method (named NM) to motion saliency estimation
to better assess the contribution of the main components of our
method. It merely consists in first computing the dominant (or global)
motion in the image. To this end, we estimate an affine motion model
with the robust multi-resolution algorithm Motion2D
\cite{Odobez1995}. No inpainting masks are extracted. The residual
flow contributing to the motion saliency map is now the difference,
over the whole image, between the computed optical flow and the
estimated parametric dominant flow.  As reported in Table
\ref{table/comparison}, we observe that the method NM yields poor
performance.  It demonstrates the importance of the flow inpainting
approach for motion saliency.

Table \ref{table/comparison} also collects comparative results of our
three variants, MSI-ns, MSI-pm and MSI-fm, with state-of-the-art
methods for saliency map estimation in videos: LGFOGR \cite{Wang2015},
SAG \cite{Wang2_2015}, RST \cite{Le2016}, STCRF \cite{Le2018} and
VSFCN \cite{Wang2018}. Results of these methods are those reported in
\cite{Le2018}, except for \cite{Wang2018}, for which we used saliency
maps provided by the authors to compute the metrics.

We carried out the experimental evaluation on the test set of DAVIS
2016, which contains 20 videos. The quantitative evaluation on the
DAVIS 2016 dataset is useful, but may generate a (small) bias. The
available ground truth on DAVIS 2016 may not fully fit the
requirements of the motion saliency task as illustrated in
Fig.~\ref{figure/visualisation results} and commented in Section
\ref{section/Qualitative evaluation}, since it is object-oriented and
binary.

For the evaluation, we use the Mean Average Error (MAE), F-Adap and
F-Max metrics, that we compute the same way as in \cite{Le2018}.  The
MAE is a pixel-wise evaluation of the saliency map $g$ compared to the
binary ground truth. F-Adap and F-Max are based on the weighted
F-Measure, in which the weight $ \beta^2 $ is set to 0.3 following
\cite{Le2018}:

\begin{equation}
  F_{\beta} = \frac{(1+\beta^2)Precision \times Recall}{\beta^2 \times Precision + Recall}
\end{equation}
F-Adap involves an adaptive threshold on each saliency map based on
the mean and standard deviation of each map, while F-Max is the
maximum of the F-Measure for thresholds varying in [0,255].

Our method MSI-ns obtains consistently satisfactory results, as it
ranks second for the three metrics. The two other variants, MSI-pm and
MSI-fm, respectively rank third and fourth, but follow MSI-ns by a
small margin. Let us recall that we obtain our results without any
learning on saliency and any appearance cues in contrast to
\cite{Le2018}, which performs the best.  Our parametric and
diffusion-based flow inpainting methods have close performance on the
DAVIS 2016 dataset. However, the latter should be more easily
generalisable, since the surrounding motion cannot be always
approximated by a single parametric motion model.

Regarding the computation time, the MSI-ns method takes 10.3s to
estimate the motion saliency map for a 854x480 frame on a 2.9 GHz
processor.  Our code is written in Python and can be further
optimised. Notably, the forward and backward streams of the workflow
could be parallelised.

\section{Conclusion}
\label{Section/conclusion}

We proposed a new paradigm to estimate motion saliency maps in video
sequences based on optical flow inpainting. It yields valued saliency
maps to highlight the presence of motion saliency in videos. We tested
our method on the DAVIS 2016 dataset, and we obtained state-of-the-art
results, while using only motion information and introducing no
learning stage. This makes our method of general
applicability. Additionally, the computed residual flow on its own
provides augmented information on motion saliency, which could be
further exploited. Our current method relies on three successive
frames. Future work will aim to further leverage the temporal
dimension by exploiting longer-term dependencies.

\vfill
\pagebreak

\bibliographystyle{IEEEbib}
\bibliography{bibliography}

\end{document}